\documentclass[runningheads, orivec]{llncs}
\usepackage[T1]{fontenc}
\usepackage{graphicx}
\usepackage{multirow}
\usepackage[normalem]{ulem}

\usepackage{color}
\usepackage{epsfig}
\usepackage{colortbl}
\usepackage[table]{xcolor}

\usepackage{hyperref}
\usepackage{times}    
\usepackage{xspace}
\usepackage[dvipsnames]{xcolor}
\usepackage{amsmath}
\usepackage{amssymb}
\usepackage{booktabs}

\usepackage{fontawesome5} 
\usepackage{academicons}  
\begin{document}
\title{Is Visual Realism Enough?\\Evaluating Gait Biometric Fidelity in Generative AI Human Animation}
\titlerunning{Is Visual Realism Enough?}
%
\author{Ivan DeAndres-Tame\inst{1}\orcidID{0009-0008-5402-1153} \and \\
Chengwei Ye\inst{2}\orcidID{0009-0009-4905-6877} \and
Ruben Tolosana\inst{1}\orcidID{0000-0002-9393-3066} \and \\
Ruben Vera-Rodriguez\inst{1}\orcidID{0000-0002-6338-8511} \and
Shiqi Yu\inst{2}\orcidID{0000-0002-5213-5877}}
\authorrunning{I. DeAndres-Tame et al.}
%
\institute{Universidad Autonoma de Madrid, Madrid, Spain \email{\{ivan.deandres,ruben.tolosana,ruben.vera\}@uam.es}\and
Southern University of Science and Technology, Shenzhen, China \email{12531329@mail.sustech.edu.cn,yusq@sustech.edu.cn}
}
\maketitle              
\begin{abstract}

Generative AI (GenAI) models have revolutionized animation, enabling the synthesis of humans and motion patterns with remarkable visual fidelity. However, generating truly realistic human animation remains a formidable challenge, where even minor inconsistencies can make a subject appear unnatural. This limitation is particularly critical when AI-generated videos are evaluated for behavioral biometrics, where subtle motion cues that define identity are easily lost or distorted. The present study investigates whether state-of-the-art GenAI human animation models can preserve the subtle spatio-temporal details needed for person identification through gait biometrics. Specifically, we evaluate four different GenAI models across two primary evaluation tasks to assess their ability to \textit{i)} restore gait patterns from reference videos under varying conditions of complexity, and \textit{ii)} transfer these gait patterns to different visual identities. Our results show that while visual quality is mostly high, biometric fidelity remains low in tasks focusing on identification, suggesting that current GenAI models struggle to disentangle identity from motion. Furthermore, through an identity transfer task, we expose a fundamental flaw in appearance-based gait recognition: when texture is disentangled from motion, identification collapses, proving current GenAI models rely on visual attributes rather than temporal dynamics.


\end{abstract}

\section{Introduction} \label{sec:intro}

Recent advances in Generative AI (GenAI) models~\cite{DenoisingDiffusionProbabilistic2020Ho,Generativeadversarialnetworks2020Goodfellow} have reshaped the landscape of digital content creation. These models can now produce highly realistic images~\cite{OmniGenUnifiedImage2025Xiao} and videos~\cite{VideoChatGPTTowards2024Maaz}, enabling the automatic synthesis of humans, environments, and motion sequences with unprecedented fidelity. Despite these achievements, the realistic animation of human motion remains one of the most challenging problems in GenAI~\cite{HumanMotionVideo2025Xue}. Human perception is highly sensitive to imperfections in the appearance and movement of other humans, triggering an instinctive rejection known as the \textit{uncanny valley}. Besides static realism, maintaining temporal coherence across frames remains a major challenge. Although individual frames can look realistic, inconsistencies in pose transitions or dynamics often lead to unnatural or incoherent motion. Addressing these issues is important both for human perception and for applications that rely on accurate motion.

\begin{figure}[t]
    \centering
    \includegraphics[width=0.9\columnwidth]{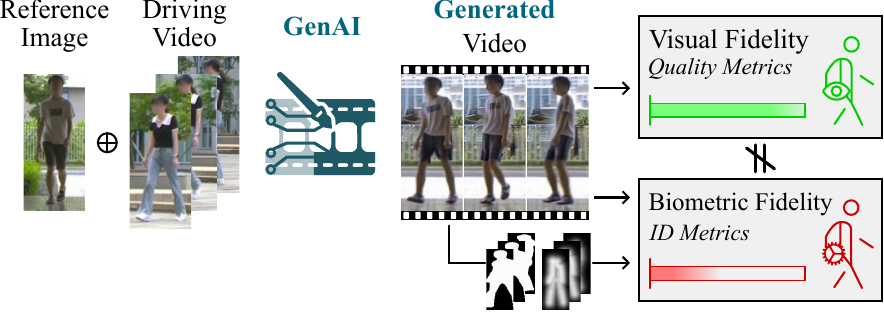}
    \caption{\textbf{Motivation.} We examine whether synthetic videos generated through state-of-the-art GenAI animation models successfully preserve the identity-specific behavioral biometric traits (gait). Both visual and biometric fidelity are analyzed in the study.}
    \label{fig:fig1}
\end{figure}

Even with these difficulties, the potential use of GenAI models extends far beyond entertainment. They can also serve as tools for synthetic data generation and model training~\cite{SystematicReviewSynthetic2024Goyal,DeepWriteSYNLineHandwriting2021Tolosana}, reducing bias~\cite{SecondFRCSynonGoing2025DeAndresTame,SDFRSyntheticData2024Shahreza}, or increasing privacy~\cite{PrivacyEnhancingFace2021Meden,OverviewPrivacyEnhancing2024Melzi}. If GenAI models could reliably replicate visual and behavioral patterns from different individuals and transfer them onto diverse visual identities, they could substantially expand existing datasets without the need for costly data acquisition~\cite{GenerationSyntheticData2024Sawicki,SyntheticDatasetsPerson2025Delussu}. Furthermore, they could enable the creation of animations that not only look realistic but also preserve the consistency and individuality of human behavior~\cite{MotioncharacterIdentitypreserving2024Fang}. These qualities allow new opportunities for simulation, behavioral analysis, and the development of AI models that rely on human behavioral traits as a source of information.

From this perspective, gait recognition represents a particularly relevant and rigorous test case. Gait is a distinctive behavioral biometric pattern characterized by temporal and spatial cues, like rhythm, stride, and posture, that uniquely identify individuals~\cite{ExploringMoreMultiple2025Jin}. This uniqueness, however, introduces a critical security dimension: the potential emergence of \textit{behavioral DeepFakes}~\cite{AttackingGaitRecognition2019Jia,IsitReally2025PedrouzoRodriguez,DeepWriteSYNLineHandwriting2021Tolosana}. If GenAI models can accurately clone these motion signatures, they could compromise biometric security systems that rely on gait, for example, for surveillance or access control. Therefore, determining whether current GenAI models can synthesize deceptive behavioral biometrics is a matter of significant safety interest. As a behavioral trait, gait represents an interesting modality for synthetic generation: realistic enough to train recognition models, yet sensitive enough to reveal the limits of motion fidelity in current GenAI models. Motivated by this, in the present study we analyze how effectively animation models can maintain identity-preserving motion traits in the context of gait synthesis and recognition. With this analysis, we aim to bridge the gap between visual realism and behavioral fidelity in AI-generated human animation, contributing to a deeper understanding of their strengths, limitations, and potential applications. The main contributions of the study are:
\begin{itemize}
    \item We examine whether state-of-the-art GenAI models can \textbf{preserve identity-specific motion traits} during human animation synthesis. We specifically analyze their ability to \textbf{reproduce gait-related behavioral information} from reference videos and transfer it to visually different identities, as illustrated in Fig~\ref{fig:fig1}.
    \item We expose a \textbf{critical limitation in current gait recognition models}: our identity transfer task reveals they work mainly as appearance-based Re-Identification models, \textbf{failing to capture temporal dynamics when texture is disentangled}.
\end{itemize}

The remainder of the paper is organized as follows. Sec.~\ref{sec:relatedworks} reviews the most relevant literature on synthetic human animation and gait analysis. Next, Sec.~\ref{sec:method} details the technical framework, covering the GenAI and gait recognition models, and evaluation metrics. Building on this, Sec.~\ref{sec:exp} outlines the experimental protocol and introduces two main evaluation tasks comprising four distinct scenarios designed to assess specific capabilities of the animation models and the used datasets. The corresponding results, including both quantitative measures and qualitative observations, are presented in Sec.~\ref{sec:results}. Finally, Sec.~\ref{sec:conclusions} summarizes the main findings and discusses future directions.

\vspace{-0.5em}

\section{Related Works}\label{sec:relatedworks}
This section first reviews state-of-the-art GenAI models for video synthesis, ranging from general video generation to specific human motion synthesis. After that, we discuss gait recognition models, which serve as the benchmarking mechanism in our study to evaluate the biometric fidelity of generated content.

\vspace{-0.5em}

\subsection{Generative Human Motion Synthesis}
Video synthesis has shifted from Generative Adversarial Networks (GANs)~\cite{Generativeadversarialnetworks2020Goodfellow} to Diffusion Models~\cite{DenoisingDiffusionProbabilistic2020Ho}, where Video Diffusion Models (VDMs) leveraging space-time U-Nets~\cite{UNetConvolutional2015Ronneberger} and latent diffusion pipelines~\cite{VideoDiffusionModels2022Ho} have emerged as the new standard for superior visual fidelity. However, applying these advancements to human animation requires precise conditioning on reference images and driving motions, a challenge where general VDMs often prioritize smoothness over physical constraints. While research on animating faces~\cite{Deepfakevideodetection2024Kaur,AudioDrivenFacial2024Jiang} has been extensively explored, analogous research for full-body motion remains scarce. Early attempts using GANs~\cite{DwnetDensewarp2019Zablotskaia,MoCoGANDecomposingMotion2018Tulyakov} focused on warping mechanisms but struggled with artifacts in complex poses and long sequences. Consequently, the field has adopted Latent Diffusion Models (LDMs), where current state-of-the-art methods are categorized by their guidance strategies. A dominant paradigm focuses on preserving high-frequency texture details by employing parallel appearance encoders combined with sparse 2D skeletal guidance~\cite{AnimateAnyoneConsistent2024Hu,ChampControllableconsistent2024Zhu}. While efficient for texture transfer, these methods try to infer body volume from simple keypoints. In contrast, alternative approaches leverage dense structural signals, such as DensePose~\cite{DensePoseDenseHuman2018Gueler}, integrated via adapter modules~\cite{MagicAnimateTemporallyConsistent2024Xu}, while others introduce 3D-aware motion tokens to better handle complex spatial dynamics beyond 2D planes~\cite{MTVCrafter4DMotion2025Ding}. By explicitly modeling the volumetric surface of the body, these techniques aim to enforce stricter structural consistency. Some architectures integrate unified noise inputs to extend coherence across longer clips~\cite{UniAnimateTamingUnified2025Wang}. Additionally, separate lines of work focus on disentangling movement from identity~\cite{IdanimatorZero2024He,MagicmeIdentity2024Ma}, although these are generally restricted to the head region and do not address full-body gait dynamics. Despite the visual plausibility achieved by these frameworks, there is limited literature analyzing their biometric fidelity, whether the generated motion preserves the unique gait signatures required for person identification.

\definecolor{metrics}{HTML}{909900}
\definecolor{synthetic}{HTML}{007a99}
\definecolor{guide}{HTML}{992200}
\definecolor{gait_input}{HTML}{8f4de5}
\definecolor{model}{HTML}{4de56b}

\vspace{-0.5em}

\subsection{Gait Recognition}
Gait recognition identifies individuals based on walking patterns, a behavioral biometric composed of cadence, stride length, and posture dynamics. Unlike other physiological traits such as face recognition, gait requires analyzing spatio-temporal data. Modern gait recognition, standardized by frameworks like OpenGait~\cite{OpenGaitRevisitingGait2023Fan}, relies on deep learning architectures generally categorized into two paradigms. \textbf{Appearance-based methods}, which operate directly on silhouettes or RGB sequences. They are robust but often heavily reliant on spatial appearance (body shape/clothing) rather than pure motion dynamics~\cite{BigGaitLearningGait2024Ye,Exploringdeepmodels2023Fan,GaitPartTemporalPart2020Fan}. Conversely,  \textbf{model-based methods} use extracted gait motion representations like 2D/3D skeletons and focus purely on structural movement evolution~\cite{SkeletonGaitGaitRecognition2024Fan,GaitgraphGraphConvolutional2021Teepe,EndEndModel2023Xu}. In this study, we use these gait recognition models as evaluators to assess the utility of GenAI models for gait recognition. By exposing AI-generated videos to these models, we can quantify whether the identity perceived by the generator matches the behavioral signature required for identification.

\vspace{-0.5em}

\section{Methodology}\label{sec:method}

\begin{figure*}[t]
    \centering
    \includegraphics[width=\textwidth]{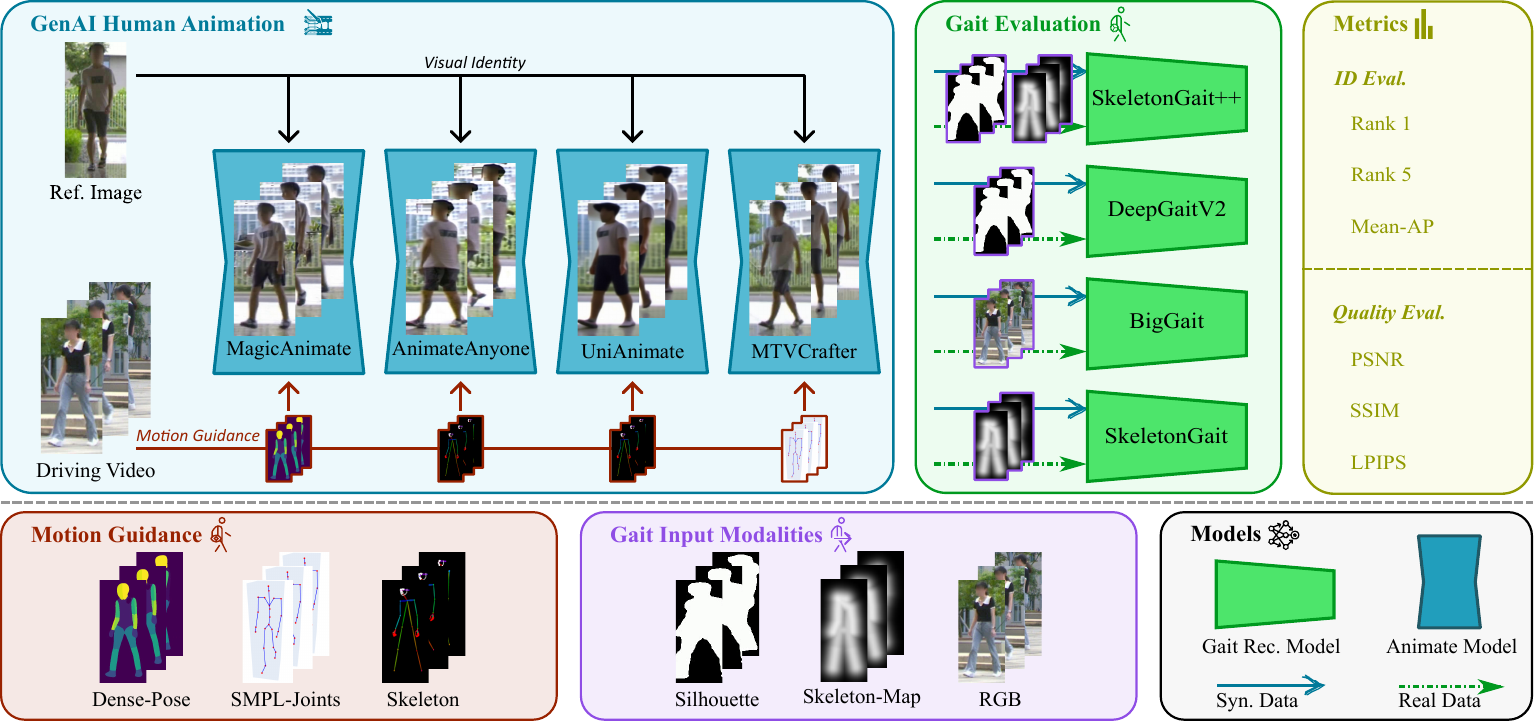}
    \caption{\textbf{Overview of the generation and evaluation pipeline.} The pipeline extracts different \textcolor{guide}{\textbf{motion guidance}} (Skeleton, SMPL, DensePose) from a driving video to condition four state-of-the-art \textcolor{synthetic}{\textbf{animation models}}. To assess biometric fidelity, we extract specific \textcolor{gait_input}{\textbf{gait modalities}} (Silhouettes, Skeleton Maps, RGB) from the generated outputs and subject them to four different \textcolor{model}{\textbf{gait recognition models}}. This allows us to quantify \textcolor{metrics}{\textbf{identity preservation and visual quality}}.}
\label{fig:fig2}
\end{figure*}

In this section, we detail how we assess the biometric fidelity of the evaluated animation models by subjecting their outputs to specialized gait recognition models. Fig.~\ref{fig:fig2} shows an overview of the generation and evaluation pipeline.

\begin{table*}[t]
\centering
\caption{\textbf{Comparison of the GenAI animation models evaluated in this study.} We highlight the motion guidance strategies and temporal mechanisms, as our results reveal that these directly influence biometric fidelity.}
\label{tab:models_comparison}
\resizebox{\textwidth}{!}{\begin{tabular}{l|c|c|c|c|c|c}
\hline
\textbf{Model} & \textbf{Motion Guidance} & \textbf{Backbone} & \textbf{\# Params.} & \textbf{Temporal Mechanism} & \textbf{Unlimited Output} & \textbf{Key Innovation} \\ \hline

\textit{MagicAnimate}~\cite{MagicAnimateTemporallyConsistent2024Xu} & DensePose & StableDiffusion & 2.5B & Dense Video ControlNet & \checkmark & Video Fusion for smoothness \\ \hline

\textit{AnimateAnyone}~\cite{AnimateAnyoneConsistent2024Hu} & 2D Skeleton & StableDiffusion & 2.4B & Temporal Attention & \checkmark & ReferenceNet for appearance \\ \hline
\textit{UniAnimate}~\cite{UniAnimateTamingUnified2025Wang} & \begin{tabular}{c}2D Skeleton+\\Ref. 2D Skeleton\end{tabular} & StableDiffusion & 1.4B & Unified Noise Input & \checkmark & Long-video coherence \\ \hline

\textit{MTVCrafter}~\cite{MTVCrafter4DMotion2025Ding} & 3D SMPL Joints & Video DiT & 7.2B & Motion-aware Attention & 49 Frames & 4D Motion Tokens (4DMoT) \\ \hline
\end{tabular}}
\end{table*}
\vspace{-0.5em}

\subsection{GenAI Human Animation}
We evaluate four state-of-the-art animation models, specifically selected to represent diverse motion guidance strategies in human synthesis: MagicAnimate~\cite{MagicAnimateTemporallyConsistent2024Xu}, AnimateAnyone~\cite{AnimateAnyoneConsistent2024Hu}, UniAnimate~\cite{UniAnimateTamingUnified2025Wang}, and MTVCrafter~\cite{MTVCrafter4DMotion2025Ding}. A comprehensive comparison of their architectural features, motion guidance strategies, and temporal mechanisms is presented in Tab.~\ref{tab:models_comparison}. All these models use a driving motion guidance, such as pose, and a reference image to generate the final animation. Although all rely on pose, the input representation differs significantly:

\begin{itemize}
    \item \textbf{UniAnimate~\cite{UniAnimateTamingUnified2025Wang}\footnote{\href{https://github.com/ali-vilab/UniAnimate}{https://github.com/ali-vilab/UniAnimate}}:} Leverages a unified video diffusion model~\cite{HighResolutionImage2022Rombach} to map the reference image, posture guidance (17 keypoints from DWPose~\cite{EffectiveWholeBody2023Yang}), and noise video into a common feature space. It introduces a unified noise input to support long-term video generation and enhance temporal coherence.

    \item \textbf{AnimateAnyone~\cite{AnimateAnyoneConsistent2024Hu}\footnote{\href{https://github.com/MooreThreads/Moore-AnimateAnyone}{https://github.com/MooreThreads/Moore-AnimateAnyone}}:} Employs a U-Net~\cite{UNetConvolutional2015Ronneberger} combined with a pose guider (skeleton images) to control motion continuity while preserving fine-grained appearance features through a custom ReferenceNet mechanism.

    \item \textbf{MagicAnimate~\cite{MagicAnimateTemporallyConsistent2024Xu}\footnote{\href{https://showlab.github.io/magicanimate}{https://showlab.github.io/magicanimate}}:} Leverages dense pose information~\cite{DensePoseDenseHuman2018Gueler} and a dense video diffusion model. It features an appearance encoder and a video fusion technique to achieve higher visual fidelity and smoothness over longer sequences compared to sparse keypoint methods.

    \item \textbf{MTVCrafter~\cite{MTVCrafter4DMotion2025Ding}\footnote{\href{https://github.com/DINGYANB/MTVCrafter}{https://github.com/DINGYANB/MTVCrafter}}:} Adopts a different approach using SMPL joints~\cite{NeuralLocalizerFields2024Sarandi} to represent the human body and models raw 3D motion sequences over time, creating 4D motion tokens (4DMoT). This mechanism aims to provide robust spatio-temporal cues by avoiding strict 2D pose alignment.
\end{itemize}

\vspace{-0.5em}

\subsection{Gait Evaluation}
To assess the preservation of behavioral biometrics, we select four representative gait recognition models widely used for benchmarking to evaluate the GenAI human animation models. These models cover the primary input modalities in gait analysis:

\begin{itemize} 
\item \textbf{DeepGaitV2~\cite{Exploringdeepmodels2023Fan}:} An established appearance-based baseline with \textit{binary silhouette sequences} as input. Its CNN-based backbone extracts features from the body shape and temporal sequence, ignoring internal texture.

\item \textbf{SkeletonGait~\cite{SkeletonGaitGaitRecognition2024Fan}:} A model relying only on \textit{skeleton sequences}. By modeling the temporal evolution of the skeleton joint coordinates, it isolates motion dynamics from appearance cues.

\item \textbf{SkeletonGait++~\cite{SkeletonGaitGaitRecognition2024Fan}:} A hybrid framework fusing \textit{skeleton-based pose information} with \textit{binary silhouettes}. This multi-modal design improves robustness by integrating structural dynamics with body shape information.

\item \textbf{BigGait~\cite{BigGaitLearningGait2024Ye}:} A comprehensive appearance-based model using \textit{RGB videos}. Leveraging DINOv2~\cite{Dinov2Learningrobust2023Oquab}, it captures rich semantic cues, separating appearance (color, clothing) from motion patterns.

\end{itemize}

In order to guarantee a fair comparison, all four models are implemented within the OpenGait framework and share a \textbf{unified feature embedding head (GaitBase)~\cite{OpenGaitRevisitingGait2023Fan}}. This architectural alignment ensures that performance variations are attributable to the quality of the generated input data across modalities, rather than discrepancies in the final feature projection or metric learning strategy.

\vspace{-0.5em}

\section{Experimental Protocol}\label{sec:exp}
\begin{figure}[t]
    \centering
    \includegraphics[width=\textwidth]{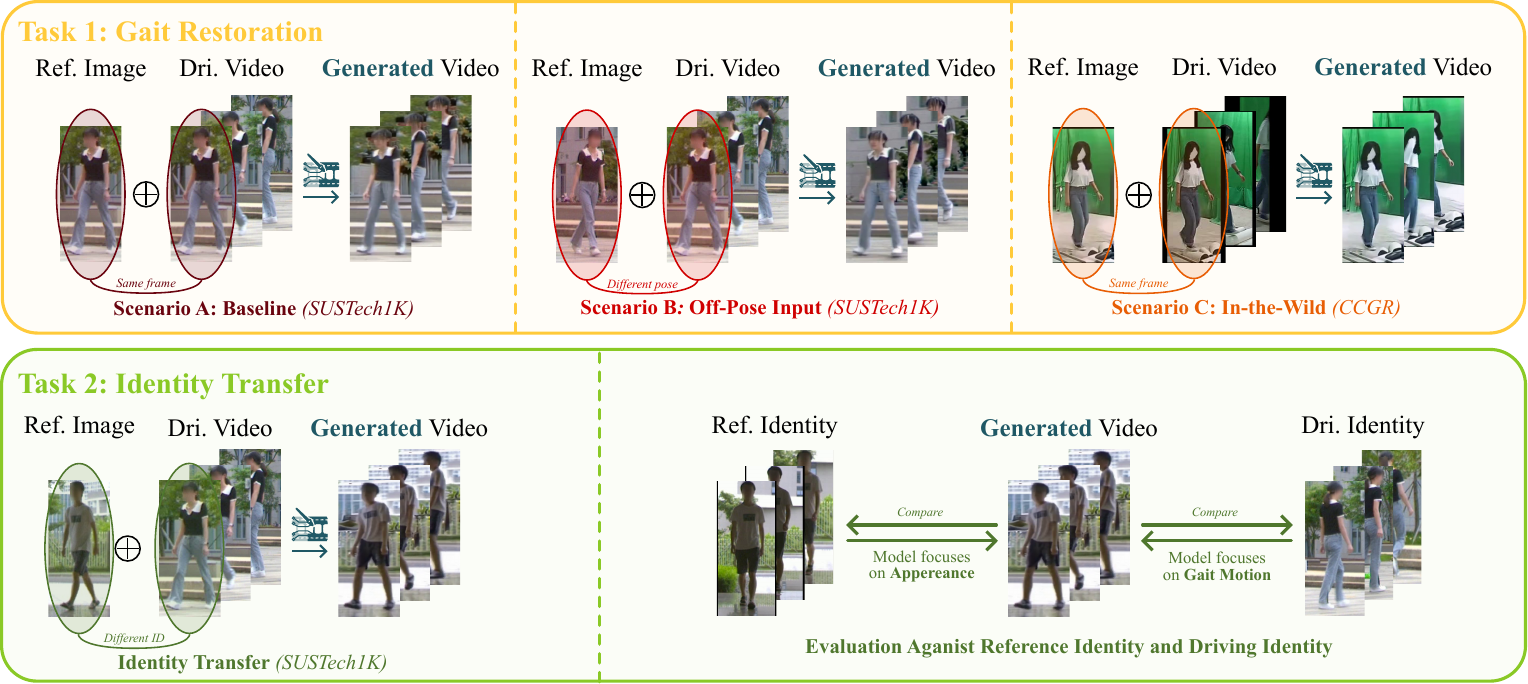}
    \caption{\textbf{Overview of the two primary evaluation tasks comprising different scenarios.} Task~1 evaluates gait restoration across (A) Baseline, (B) Off-Pose, and (C) In-the-Wild scenarios. Task~2 examines zero-shot identity transfer. The Evaluation Method block details the dual validation protocol against reference (appearance) and driving (motion) identities.}
    \label{fig:fig3}
\end{figure}

This section describes the databases and the experimental protocol designed to rigorously evaluate the reconstruction fidelity, robustness, and identity preservation capabilities of the generated gait animations. The overall experimental structure, detailing the input variations for each task, is summarized in Fig.~\ref{fig:fig3}. We divide the evaluation into two main tasks: one focused on \textbf{Gait Restoration}, which comprises three scenarios allowing for Ground-Truth comparison, and another focused on \textbf{Identity Transfer}, designed to test the disentanglement of identity from motion under strict zero-shot conditions.

\vspace{-0.5em}

\subsection{Gait Recognition Databases}
In order to ensure the robustness and generalizability of our evaluation, we use two different gait recognition datasets: \textbf{SUSTech1K~\cite{LidarGaitBenchmarking3D2023Shen}}, a large-scale dataset focusing on controlled environments with various viewing angles, typically used for general gait performance benchmarking, and \textbf{CCGR-MINI (Cross-Condition Gait Recognition)~\cite{CrossCovariateGait2024Zou}}, a dataset specifically designed to challenge models under extreme cross-view and cross-condition scenarios, such as changes in carrying objects or clothing.

To ensure a robust foundation, we select $250$ identities from each dataset, generating four videos per identity for each GenAI model, resulting in $16,000$ generated videos ($1,000$ per GenAI method and dataset). Crucially, none of these test identities or sequences are included in the training datasets of the evaluated animation models. Consequently, all evaluations are performed on strictly unseen data.

\vspace{-0.5em}

\subsection{Tasks}

\vspace{-0.5em}
\noindent\textbf{Task~1: Gait Restoration.}
This task evaluates the model's ability to reconstruct motion. We group three scenarios under this task to measure fidelity across increasing levels of difficulty. First, in \textit{Scenario A (Baseline)}, we establish a benchmark for visual fidelity and identity preservation under perfect alignment by animating the first frame of a video using the remaining frames as the driving motion. Second, in \textit{Scenario B (Off-Pose Input)}, we introduce a pose misalignment between the reference image and the driving motion to test the resilience against initialization noise. Comparison with the baseline quantifies the sensitivity of each method to source–driving pose discrepancies. Finally, \textit{Scenario C (In-the-Wild)} evaluates real-world performance using the challenging CCGR-MINI dataset by, again, animating the first frame of the video using the remaining as the driving motion. This assesses the ability to maintain fidelity under complex conditions, such as non-standard camera angles and dynamic movements, providing insight into practical robustness outside controlled environments.

\vspace{5pt}

\noindent\textbf{Task~2: Identity Transfer (Zero-Shot).}
In this task, we test the \textit{disentanglement of identity from motion} under zero-shot conditions. We use a reference image of one identity to animate a driving motion of a different identity. Since the model has never seen the reference identity performing this specific motion pattern, this scenario represents the most challenging test in terms of generalization. With no visual ground truth available, the evaluation focuses on the transferability of gait features. We employ a dual validation protocol: we compare the generated output against the \textit{Reference Identity} to assess appearance preservation, and against the \textit{Driving Identity} to quantify the fidelity of the transferred gait motion.

\vspace{-0.5em}

\subsection{Metrics}\label{subsec:metrics}
In order to perform a comprehensive evaluation, we employ both visual quality and identity-based recognition metrics.

\vspace{5pt}

\noindent{\textbf{Visual Quality Metrics.}}
These metrics are exclusively applied to \textit{Task~1 (Gait Restoration)}, as Task~2 (Identity Transfer) generates novel views without a pixel-aligned ground truth. Crucially, to focus the evaluation strictly on the generated subject and mitigate the impact of background hallucinations, we apply a segmentation mask to both the generated and ground-truth frames. This ensures that all pixel-wise and perceptual comparisons are restricted to the person foreground.
\begin{itemize}
    \item\textbf{SSIM (Structural Similarity Index)~\cite{Imagequalityassessment2004Wang}:} Measures the perceptual similarity between the generated and ground-truth videos by comparing luminance, contrast, and structural information.
    \item\textbf{PSNR (Peak Signal-to-Noise Ratio)~\cite{Imagequalitymetrics2010Hore}:} A classical pixel-wise measure of reconstruction fidelity based on Mean Squared Error (MSE).
    \item\textbf{LPIPS (Learned Perceptual Image Patch Similarity)~\cite{UnreasonableEffectivenessDeep2018Zhang}:} Computes perceptual distances using deep feature representations, correlating more strongly with human perception of visual quality.
\end{itemize}

\vspace{2pt}

\noindent{\textbf{Identity and Gait Recognition Metrics.}}
Using features extracted by BigGait, DeepGaitV2, SkeletonGait++, and SkeletonGait, we compute \textbf{Rank-1\&5 Accuracy} to measure retrieval success, and \textbf{Mean Average Precision (mAP)} to evaluate overall ranking performance across recall levels.

\vspace{-0.5em}

\section{Results} \label{sec:results}

This section reports the results of our experiments. We first evaluate the visual fidelity of the generated sequences using standard image quality and restoration metrics. Next, we perform a comprehensive analysis of the biometric fidelity, evaluating the ability of the GenAI human animation models to preserve the identity-specific gait motions required for person identification across the four proposed experiments. 

\begin{table}[t]
\centering
\caption{\textbf{Visual Reconstruction Metrics.} Comparison of structural (SSIM) and pixel-level (PSNR, LPIPS) fidelity. Higher SSIM/PSNR and lower LPIPS indicate better performance. \\ \textit{Note:} Task 2 is excluded as it lacks pixel-aligned ground truth.}
\label{tab:vquality}
\resizebox{\linewidth}{!}{  
\begin{tabular}{l|ccc|ccc|ccc}
\hline
\multirow{2}{*}{\textbf{Model}} & \multicolumn{3}{c|}{\textbf{SUSTech1K} \textit{(Baseline)}} & \multicolumn{3}{c|}{\textbf{SUSTech1K} \textit{(Noise)}} & \multicolumn{3}{c}{\textbf{CCGR-MINI} \textit{(In-the-Wild)}} \\ \cline{2-10} 
 & \textbf{SSIM$\uparrow$} & \textbf{PSNR$\uparrow$} & \textbf{LPIPS$\downarrow$} & \textbf{SSIM$\uparrow$} & \textbf{PSNR$\uparrow$} & \textbf{LPIPS$\downarrow$} & \textbf{SSIM$\uparrow$} & \textbf{PSNR$\uparrow$} & \textbf{LPIPS$\downarrow$} \\ \hline
MagicAnimate & \textbf{0.82} & \textbf{19.51} & \textbf{0.17} & \textbf{0.82} & \textbf{19.13} & \textbf{0.18} & 0.72 & 15.41 & 0.28 \\
AnimateAnyone & 0.77 & 16.50 & 0.26 & 0.77 & 16.46 & 0.26 & 0.70 & 15.15 & 0.31 \\
UniAnimate & 0.80 & 17.73 & 0.22 & 0.80 & 17.89 & 0.21 & \textbf{0.74} & \textbf{16.02} & \textbf{0.28} \\
MTVCrafter & 0.81 & 18.18 & 0.20 & 0.79 & 17.14 & 0.23 & 0.71 & 15.16 & 0.32 \\ \hline
\end{tabular}
}
\end{table}

\begin{figure}[t]
    \centering
    \includegraphics[width=\textwidth]{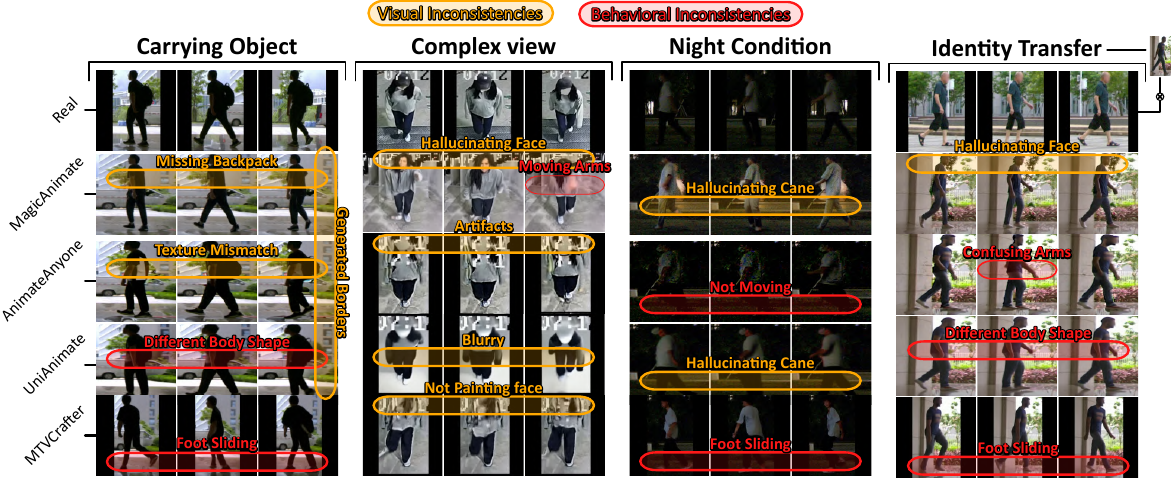}
    \caption{\textbf{Failures produced by different GenAI models across different scenarios.} We visualize the specific artifacts that degrade biometric fidelity. We highlight the \textcolor{orange}{\textbf{visual inconsistencies}}, and the \textcolor{red}{\textbf{behavioral inconsistencies}}.}
    \label{fig:fig4}
\end{figure}

\vspace{-0.5em}
\subsection{Visual Reconstruction Quality}

Tab.~\ref{tab:vquality} summarizes the pixel-level and perceptual quality of the generated videos across the first task. Note that the \textit{Identity Transfer} task is excluded from this comparison as it lacks a ground truth for comparison. In general, MagicAnimate shows superior performance on the standard SUSTech1K dataset, achieving in the baseline case the highest SSIM ($0.82$) and PSNR ($19.51$) and the lowest LPIPS ($0.17$). This suggests that its dense video diffusion prior is highly effective at maintaining structural consistency and sharpness in controlled environments. However, on the more challenging CCGR-MINI dataset, UniAnimate achieves the best performance ($0.74$ SSIM, $16.02$ PSNR, $0.28$ LPIPS). The unified noise input strategy of UniAnimate appears significantly more robust to complex background variations and camera shifts than the dense-pose approach. Notably, AnimateAnyone consistently scores lower on these pixel-aligned metrics across all subsets (\textit{e.g.}, SSIM $0.77$ on SUSTech1K), despite its visual popularity. As we will see in the next section, this creates an interesting contrast: a model can have lower pixel-perfect alignment (lower SSIM) but superior texture preservation for identification (higher RGB Rank-1). To further investigate these failure cases, Fig.~\ref{fig:fig4} visualizes generation artifacts, categorized into visual and behavioral inconsistencies. In terms of appearance, we observe significant semantic failures like hallucinating accessories, faces, or limbs. Crucially, we identify a systematic structural bias: most models, with the notable exception of MagicAnimate, exhibit unnatural volumetric thickening, rendering subjects significantly wider than the ground truth. This aspect ratio distortion suggests that sparse skeleton guidance fails to constrain the volumetric boundaries of the subject during texture warping.

\begin{figure*}[t]
    \centering
    \includegraphics[width=\textwidth]{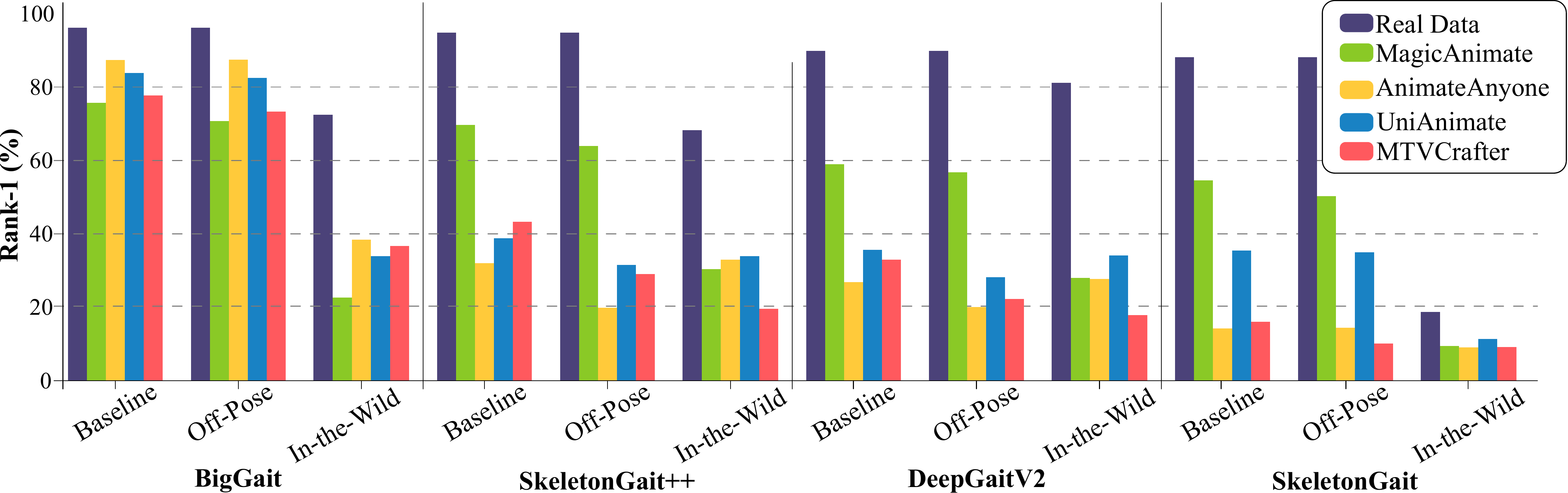}
    \caption{\textbf{Task~1: Gait Restoration Biometric Performance.} The comparison highlights a modality gap: AnimateAnyone excel in texture preservation (BigGait), while MagicAnimate dominates in motion fidelity (SkeletonGait). Performance consistently degrades in the unconstrained In-the-Wild scenario.}
    \label{fig:fig5}
\end{figure*}

\vspace{-0.5em}
\subsection{Biometric Performance for Task 1: Gait Restoration}

\begin{table}[t]
\centering
\caption{\textbf{Task~1: Gait Restoration Results.} Quantitative evaluation of biometric fidelity across three increasing levels of difficulty: Baseline, Robustness to Pose-Offset, and Real-World Generalization. The best result per block is \textbf{bolded} and Real Data is \underline{underlined}. All the results are presented as percentages (\%).}
\label{tab:task1_restoration}
\renewcommand{\arraystretch}{1.0} 
\setlength{\tabcolsep}{3.5pt}     
\resizebox{\textwidth}{!}{
\begin{tabular}{c|c|ccc|ccc|ccc}
\hline
\multirow{2}{*}{\textbf{Gait Model}} & \multirow{2}{*}{\textbf{Data Source}} & \multicolumn{3}{c|}{\textbf{Scenario A: Baseline}} & \multicolumn{3}{c|}{\textbf{Scenario B: Noise}} & \multicolumn{3}{c}{\textbf{Scenario C: Wild}} \\
 &  & \textbf{R-1}$\uparrow$ & \textbf{R-5}$\uparrow$ & \textbf{mAP}$\uparrow$ & \textbf{R-1}$\uparrow$ & \textbf{R-5}$\uparrow$ & \textbf{mAP}$\uparrow$ & \textbf{R-1}$\uparrow$ & \textbf{R-5}$\uparrow$ & \textbf{mAP}$\uparrow$ \\ \hline \hline

\multirow{5}{*}{\begin{tabular}[c]{@{}c@{}}BigGait\\ (RGB)\end{tabular}} 
 & \underline{Real Data} & \underline{96.30} & \underline{99.90} & \underline{92.54} & \underline{96.30} & \underline{99.90} & \underline{92.54} & \underline{22.60} & \underline{42.70} & \underline{21.56} \\
 & MagicAnimate & 75.80 & 88.20 & 71.64 & 70.90 & 87.00 & 67.92 & 7.10 & 19.30 & 9.29 \\
 & \textbf{AnimateAnyone} & \textbf{87.50} & \textbf{97.60} & \textbf{84.30} & \textbf{87.60} & \textbf{97.20} & \textbf{85.03} & \textbf{12.00} & \textbf{29.90} & \textbf{14.25} \\
 & UniAnimate & 84.00 & 94.00 & 76.23 & 82.60 & 93.20 & 76.85 & 10.60 & 25.90 & 12.09 \\
 & MTVCrafter & 77.80 & 92.84 & 71.77 & 73.46 & 91.42 & 68.75 & 11.48 & 26.99 & 13.13 \\ \hline
 
\multirow{5}{*}{\begin{tabular}[c]{@{}c@{}}SkeletonGait++ \\ (Skeleton+Sil)\end{tabular}} 
 & \underline{Real Data} & \underline{95.00} & \underline{99.60} & \underline{82.01} & \underline{95.00} & \underline{99.60} & \underline{82.01} & \underline{21.30} & \underline{38.90} & \underline{19.74} \\
 & \textbf{MagicAnimate} & \textbf{69.80} & \textbf{85.90} & \textbf{58.17} & \textbf{64.06} & \textbf{84.68} & \textbf{54.98} & 9.50 & 23.60 & 10.84 \\
 & AnimateAnyone & 32.10 & 57.70 & 24.38 & 20.00 & 20.00 & 13.12 & 10.31 & 24.32 & 11.49 \\ 
 & UniAnimate & 38.90 & 64.00 & 29.71 & 31.66 & 55.51 & 24.96 & \textbf{10.61} & \textbf{24.62} & \textbf{11.86} \\ 
 & MTVCrafter & 43.39 & 66.30 & 38.82 & 29.16 & 49.95 & 27.94 & 6.14 & 17.52 & 8.24 \\ \hline
 
\multirow{5}{*}{\begin{tabular}[c]{@{}c@{}}DeepGaitV2\\ (Silhouette)\end{tabular}} 
 & \underline{Real Data} & \underline{89.80} & \underline{97.70} & \underline{74.93} & \underline{89.80} & \underline{97.70} & \underline{74.93} & \underline{20.50} & \underline{38.70} & \underline{20.22} \\
 & \textbf{MagicAnimate} & \textbf{58.90} & \textbf{79.40} & \textbf{49.88} & \textbf{56.70} & \textbf{78.90} & \textbf{48.24} & 8.70 & 22.20 & 11.20 \\
 & AnimateAnyone & 26.70 & 50.70 & 21.29 & 20.00 & 30.00 & 12.75 & 8.60 & 21.10 & 10.31 \\ 
 & UniAnimate & 35.50 & 60.20 & 27.40 & 28.13 & 50.75 & 23.89 & \textbf{10.60} & \textbf{23.60} & \textbf{12.19} \\
 & MTVCrafter & 32.90 & 56.00 & 30.48 & 22.20 & 41.57 & 22.54 & 5.54 & 14.10 & 7.43 \\ \hline

\multirow{5}{*}{\begin{tabular}[c]{@{}c@{}}SkeletonGait\\ (Skeleton)\end{tabular}} 
 & \underline{Real Data} & \underline{88.00} & \underline{96.70} & \underline{73.78} & \underline{88.00} & \underline{96.70} & \underline{73.78} & \underline{5.80} & \underline{16.90} & \underline{7.05} \\
 & \textbf{MagicAnimate} & \textbf{54.51} & \textbf{77.66} & \textbf{44.44} & \textbf{50.15} & \textbf{71.17} & \textbf{40.90} & 2.90 & 8.80 & 4.38 \\
 & AnimateAnyone & 14.10 & 33.80 & 12.60 & 14.29 & 22.86 & 12.49 & 2.80 & 9.31 & 4.43 \\ 
 & UniAnimate & 35.40 & 58.40 & 29.65 & 34.83 & 57.76 & 30.72 & \textbf{3.50} & \textbf{11.31} & \textbf{5.11} \\
 & MTVCrafter & 15.94 & 36.23 & 15.59 & 9.99 & 27.45 & 11.24 & 2.82 & 6.65 & 3.78 \\ \hline

\end{tabular}
}
\end{table}

\begin{figure*}[t]
    \centering
    \includegraphics[width=0.9\textwidth]{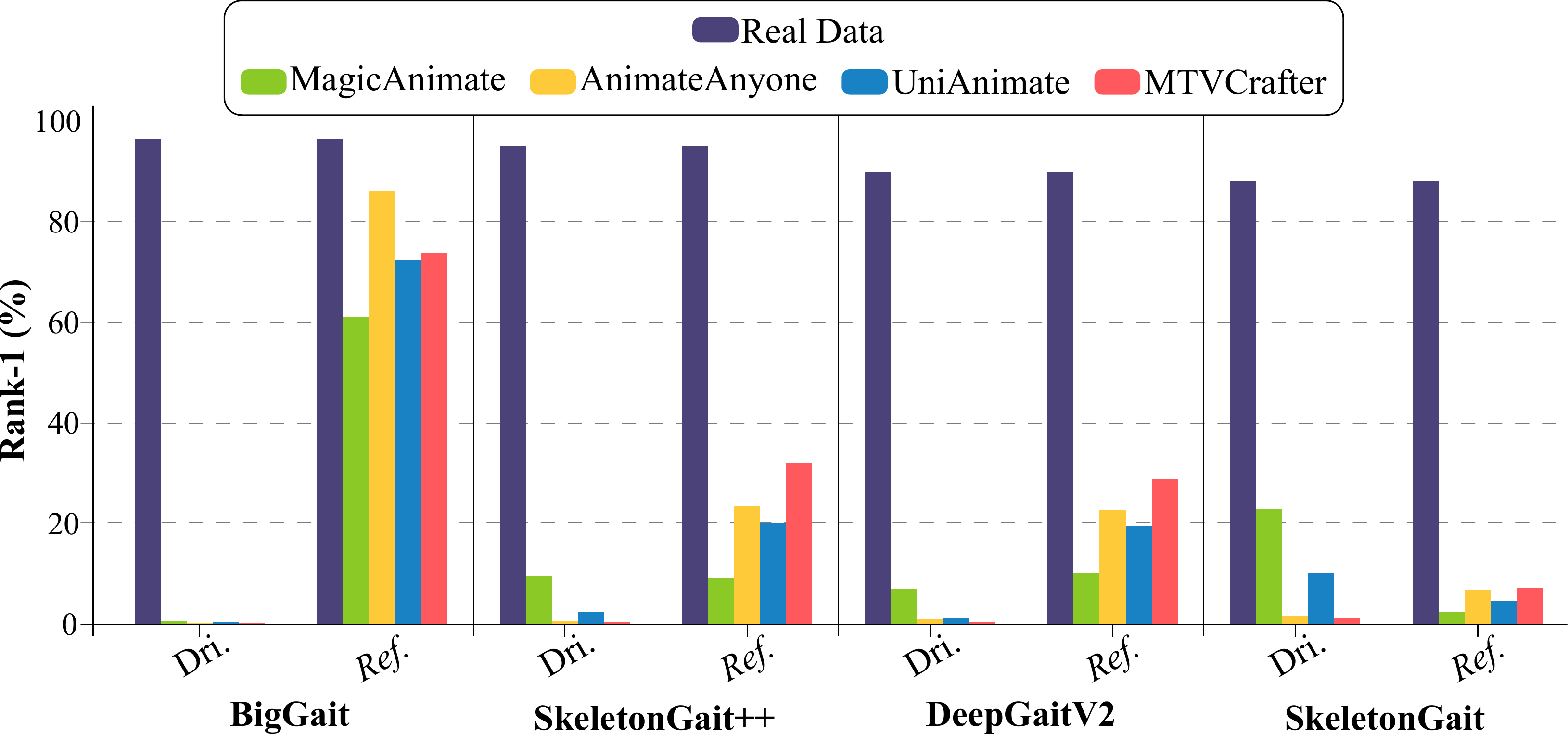}
    \caption{\textbf{Task~2: Identity Transfer Biometric Performance.} We evaluate identity fidelity against the Driving Motion (Dri.) versus the Reference Image (Ref.). The collapse in Dri. accuracy proves that current models fail to transfer motion dynamics, while high Ref. scores indicate an over-reliance on visual appearance (Re-ID).}
    \label{fig:fig6}
\end{figure*}

Fig.~\ref{fig:fig5} illustrates the biometric fidelity for Task~1 (Gait Restoration) across the three progressively complex scenarios: base standard animation, offset-pose initialization, and unconstrained in-the-wild environments. A clear modality gap is evident: in the RGB domain (BigGait), skeleton-guided models like AnimateAnyone and UniAnimate closely match the real data in the Baseline and Off-Pose scenarios. However, this trend reverses for structural evaluators (SkeletonGait++, DeepGaitV2, and SkeletonGait), where MagicAnimate consistently outperforms competitors, confirming its superior kinematic fidelity. Additionally, the uniform performance drop observed in the In-the-Wild scenario reflects the complexity of generating coherent motion under unconstrained conditions. We observe a systematic performance decay: as condition complexity increases (from pose misalignment to wild camera movements), the advantage of dense-pose guidance (MagicAnimate) narrows, eventually collapsing in favor of unified noise strategies (UniAnimate). Tab.~\ref{tab:task1_restoration} details the identification performance across these scenarios.

In the \textit{baseline scenario}, a comprehensive analysis reveals a significant gap between visual realism and behavioral patterns. In the RGB domain (BigGait), skeleton-guided models demonstrate superior texture preservation. Specifically, AnimateAnyone leads with a Rank-1 of $87.50\%$, closely followed by UniAnimate ($84.00\%$), both approaching the Real Data baseline ($96.30\%$). However, when evaluated on the DeepGaitV2 model (binary silhouettes), both models suffer a drastic drop: AnimateAnyone falls to $26.70\%$ and UniAnimate to $35.50\%$. In contrast, MagicAnimate achieves $58.90\%$. This gap indicates that while methods like AnimateAnyone effectively copy clothing texture, they fail to generate the consistent body shape required for silhouette-based identification. This finding is reinforced by evaluating with SkeletonGait++ and SkeletonGait, where MagicAnimate leads with $69.80\%$ and $54.51\%$ respectively, confirming that dense-pose guidance preserves underlying kinematics and volumetric consistency significantly better than sparse skeleton guidance.

The introduction of \textit{offset-pose initialization} further penalizes this structural fidelity. While RGB performance (BigGait) remains stable across all models (varying only $\sim2$-$5$\%), structural fragility becomes evident. Notably, MTVCrafter exhibits significant sensitivity, as its accuracy on pure SkeletonGait drops from $15.94\%$ in the baseline scenario to $9.99\%$ with offset initialization, suggesting its 3D-aware token strategy is highly susceptible to initialization noise. Conversely, UniAnimate shows relative stability (dropping slightly from $35.40\%$ to $34.83\%$), although it remains far below the structural robustness of MagicAnimate, which retains the highest accuracy across DeepGaitV2 ($56.70\%$), SkeletonGait ($50.15\%$), and SkeletonGait++ ($64.06\%$).

The performance of the generators drop significantly in the \textit{in-the-wild scenario}. First, it is crucial to note that extreme viewing angles and dynamic camera movements degrade the Real Data baseline (\textit{e.g.}, from $\sim95\%$ to $21.30\%$ on SkeletonGait++ and $22.60\%$ on BigGait). Under these conditions, the performance hierarchy shifts dramatically. UniAnimate emerges as the most robust model for structural preservation, achieving the highest Rank-1 on SkeletonGait++ ($10.61\%$). This suggests that its unified noise input strategy offers superior stability against camera shifts compared to MagicAnimate, whose performance collapses (from leader in the baseline scenario to $9.50\%$) likely due to DensePose estimation errors in the wild. Conversely, AnimateAnyone keeps dominance in the RGB domain ($12.00\%$ on BigGait). However, it is crucial to acknowledge that the absolute performance across all methods is low (peaking at only $\sim12\%$ Rank-1). This sharp decline compared to the baseline scenario highlights a significant gap in current generative capabilities: while state-of-the-art models handle simple backgrounds and still camera settings well, they struggle to preserve identity-preserving motion and shape features under complex, in-the-wild camera dynamics.

\vspace{-0.5em}
\subsection{Biometric Performance for Task~2: Identity Transfer}

Fig.~\ref{fig:fig6} visualizes the disentanglement analysis by comparing identification accuracy against the Driving Motion (Dri.) versus the Reference Image (Ref.). The results expose a fundamental limitation: while models like AnimateAnyone achieve high scores on the Reference metric (indicating strong texture preservation), they suffer a near-total collapse when evaluated against the Driving motion. This confirms that they operate primarily as appearance-based Re-Identification models. Notably, MagicAnimate on SkeletonGait is the only case that preserves motion dynamics to a significant extent when visual attributes are disentangled. Tab.~\ref{tab:exp3_combined_final} details the identification performance in this task.

When evaluating identification against the driving subject, appearance-based evaluators naturally collapse due to the visual domain shift. For instance, on the BigGait (RGB) model, performance drops to near-zero (\textit{e.g.}, AnimateAnyone at $0.20\%$). Theoretically, effective motion transfer should yield high recognition rates on skeleton-based models, which ignore texture. However, most generative models fail here as well ($\sim0-2\%$). Notably, MagicAnimate achieves a Rank-1 of $22.72\%$ on SkeletonGait, proving that dense-pose guidance successfully transfers the underlying gait motion even when the visual texture is completely replaced.

In order to accurately assess the extent to which these models can disentangle kinematics from visual attributes, the right side of Tab.~\ref{tab:exp3_combined_final} analyzes identification accuracy against the Reference identity. Here, we observe a complete inversion of performance. On the BigGait (RGB) model, AnimateAnyone achieves an impressive $86.10\%$ match with the Reference identity, comparable to its performance in the gait restoration baseline ($87.50\%$), but fails to match the Driving motion ($0.40\%$). Similarly, MTVCrafter shows strong appearance retention on shape-aware models (scoring $31.99\%$ on SkeletonGait++ vs Ref), yet fails to capture the driving dynamics ($0.40\%$). This contrast leads to a critical finding regarding current gait recognition models. The fact that AnimateAnyone dominates the RGB benchmarks on the Reference identity while failing the motion benchmarks confirms that state-of-the-art RGB gait recognition models primarily work as Re-Identification (Re-ID) models, over-relying on static texture and silhouette shape rather than temporal dynamics. Conversely, the specific success of MagicAnimate on the Driving/Skeleton metrics ($22.72\%$ for Dri. vs. $2.30\%$ for Ref.) demonstrates true motion transfer, confirming that extracting explicit gait modalities is the only viable path to recognize pure gait patterns free from appearance-based noise.

\begin{table*}[t]
\centering
\caption{\textbf{Task~2:} Comprehensive Identity Transfer Results. We compare the identification performance based on the \textit{Driving Motion} versus the \textit{Reference Image}. The best result per block is \textbf{bolded} and Real Data is \underline{underlined}. All the results are presented as percentages (\%).}
\label{tab:exp3_combined_final}
\renewcommand{\arraystretch}{1.0} 
\setlength{\tabcolsep}{3.5pt}
\resizebox{0.75\textwidth}{!}{
\begin{tabular}{c|c|ccc||ccc}
\hline
\multirow{2}{*}{\textbf{Gait Model}} & \multirow{2}{*}{\textbf{Data Source}} & \multicolumn{3}{c||}{\textbf{Driving Motion}} & \multicolumn{3}{c}{\textbf{Reference Image}} \\ \cline{3-8} 
 &  & \textbf{R-1$\uparrow$} & \textbf{R-5$\uparrow$} & \textbf{mAP$\uparrow$} & \textbf{R-1$\uparrow$} & \textbf{R-5$\uparrow$} & \textbf{mAP$\uparrow$} \\ \hline
\hline

\multirow{5}{*}{\begin{tabular}[c]{@{}c@{}}BigGait\\ (RGB)\end{tabular}} 
 & \underline{Real Data} & \underline{96.30} & \underline{99.90} & \underline{92.54} & \underline{96.30} & \underline{99.90} & \underline{92.54} \\
 & MagicAnimate & \textbf{0.60} & \textbf{1.50} & \textbf{1.44} & 61.00 & 78.00 & 58.63 \\
 & AnimateAnyone & 0.20 & 0.90 & 1.21 & \textbf{86.10} & \textbf{97.10} & \textbf{81.89} \\
 & UniAnimate & 0.40 & 1.20 & 1.48 & 72.20 & 82.90 & 70.02 \\
 & MTVCrafter & 0.20 & 1.01 & 1.27 & 73.66 & 90.01 & 69.07 \\ \hline
 
\multirow{5}{*}{\begin{tabular}[c]{@{}c@{}}SkeletonGait++ \\ (Skeleton+Sil)\end{tabular}} 
 & \underline{Real Data} & \underline{95.00} & \underline{99.60} & \underline{82.01} & \underline{95.00} & \underline{99.60} & \underline{82.01} \\
 & MagicAnimate & \textbf{9.41} & \textbf{21.02} & \textbf{10.15} & 9.11 & 23.92 & 11.27 \\
 & AnimateAnyone & 0.60 & 1.80 & 1.86 & 23.32 & 47.45 & 19.82 \\
 & UniAnimate & 2.30 & 5.50 & 3.56 & 20.10 & 39.50 & 16.69 \\
 & MTVCrafter & 0.40 & 1.72 & 1.49 & \textbf{31.99} & \textbf{56.91} & \textbf{30.67} \\ \hline

\multirow{5}{*}{\begin{tabular}[c]{@{}c@{}}DeepGaitV2\\ (Silhouette)\end{tabular}} 
 & \underline{Real Data} & \underline{89.80} & \underline{97.70} & \underline{74.93} & \underline{89.80} & \underline{97.70} & \underline{74.93} \\
 & MagicAnimate & \textbf{6.90} & \textbf{15.70} & \textbf{7.86} & 10.00 & 20.70 & 11.09 \\
 & AnimateAnyone & 0.90 & 1.90 & 1.64 & 22.60 & 45.70 & 18.61 \\
 & UniAnimate & 1.10 & 3.50 & 2.53 & 19.40 & 40.60 & 17.23 \\
 & MTVCrafter & 0.40 & 2.12 & 1.26 & \textbf{28.76} & \textbf{48.03} & \textbf{25.46} \\ \hline

\multirow{5}{*}{\begin{tabular}[c]{@{}c@{}}SkeletonGait\\ (Skeleton)\end{tabular}} 
 & \underline{Real Data} & \underline{88.00} & \underline{96.70} & \underline{73.78} & \underline{88.00} & \underline{96.70} & \underline{73.78} \\
 & MagicAnimate & \textbf{22.72} & \textbf{42.94} & \textbf{19.82} & 2.30 & 6.61 & 3.66 \\
 & AnimateAnyone & 1.60 & 4.70 & 2.13 & 6.80 & 18.10 & 7.46 \\
 & UniAnimate & 10.00 & 19.90 & 10.37 & 4.60 & 12.90 & 5.21 \\
 & MTVCrafter & 1.01 & 3.63 & 1.92 & \textbf{7.16} & \textbf{22.70} & \textbf{9.82} \\ \hline
\end{tabular}
}
\end{table*}

\vspace{-0.5em}

\section{Conclusion}
\label{sec:conclusions}
This study has presented a framework to evaluate the biometric fidelity of GenAI human animation models, revealing three critical insights. First, we demonstrate a \textbf{discrepancy between visual realism and behavioral fidelity}: models prioritizing texture (\textit{e.g.}, AnimateAnyone) often mask kinematic inconsistencies, while structure-aware models (MagicAnimate) excel in controlled settings but lack robustness under in-the-wild scenarios. Second, no single GenAI model is universal; performance is highly domain-dependent, with UniAnimate offering superior stability under complex camera views (CCGR-MINI dataset). Third, and most critically, we expose a \textbf{fundamental flaw in gait recognition}. Crucially, our results show a divergence between restoration and transfer: models like AnimateAnyone excel at ``copy-pasting'' texture in Task~1 (Restoration) but fail to emulate motion in Task~2 (Transfer). The collapse of appearance-based evaluators in our zero-shot task proves they operate primarily as static Re-Identification models, failing to capture temporal dynamics when texture is disentangled.

\noindent \textbf{Future Work.} To bridge this gap, future research must move beyond frame-wise quality, integrating \textbf{biometric-aware objectives}, such as temporal skeleton consistency losses, into diffusion training. This is essential to ensure that the generated motion is not just visually plausible, but behaviorally authentic and secure against DeepFakes.

\vspace{-1em}
\subsubsection{Acknowledgements} This project has been supported by PowerAI+ (SI4/PJI/2024-00062 Comunidad de Madrid and UAM), Cátedra ENIA UAM-Veridas en IA Responsable (NextGenerationEU PRTR TSI-100927-2023-2), Becas Santander Investigación | Predoctoral, and National Natural Science Foundation of China (Grant 62476120).

%
%
%
\bibliographystyle{splncs04}
\bibliography{ref}
\end{document}